# Multi-modal Aggregation for Video Classification

Chen Chen
Alibaba Group iDST
chenen.cc@alibaba-inc.com

Xiaowei Zhao
Alibaba Group iDST
zhiquan.zxw@alibaba-inc.com

Yang Liu
Alibaba Group iDST
panjun.ly@alibaba-inc.com

## ABSTRACT

In this paper, we present a solution to Large-Scale Video Classification Challenge (LSVC2017) [1] that ranked the 1st place. We focused on a variety of modalities that cover visual, motion and audio. Also, we visualized the aggregation process to better understand how each modality takes effect. Among the extracted modalities, we found Temporal-Spatial features calculated by 3D convolution quite promising that greatly improved the performance. We attained the official metric mAP 0.8741 on the testing set with the ensemble model.

## 1 INTRODUCTION

Video classification is a challenging task in computer vision that has significant attention in recent years along with more and more large-scale video datasets. Compared with image classification, video classification needs to aggregate frame level features to video level knowledge. More modalities can be extracted in videos like audio, motion, ASR etc. Multi-modalities are mutual complement to each other in most cases.

The recent competition entitled "Large-Scale Video Classification Challenge" provides a platform to explore new approaches for realistic setting video classification. The dataset [2] contains over 8000 hours with 500 categories which cover a range of topics like social events, procedural events, objects, scenes, etc. The training/validation/test set has 62000/15000/78000 untrimmed videos respectively. The evaluation metric is mean Average Precision (mAP) across all categories. The organizers provide frame level features with 1fps based on VGG. They also give raw videos for the whole dataset and participants are allowed to extract any modality.

## 2 APPROACH

### 2.1 Video Classification Architecture

For the video classification method, the first step is to extract frame level CNNs activations as intermediate features. And then aggregate the features through pooling layers like VLAD, Bag-of-visual-words, LSTM and GRU. In previous YouTube-8M competition [3], the frame level features were restricted to officially provided ImageNet pre-trained inception v3 activation thus the participants can only focus on aggregation methods. However, in LSVC2017 competition, since the raw videos are provided and the dataset scale is suitable, we put emphasis on modality extraction and used VLAD as the aggregation layer. Figure 1 shows our architecture for multi-modal aggregation for video classification.

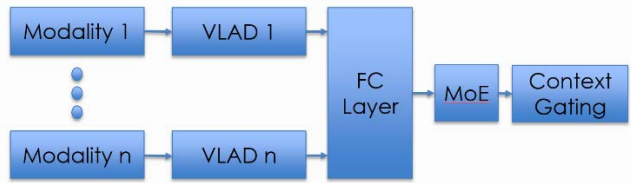

**Figure 1:** *Overview of the video classification architecture.*

*2.1.1 Modality Extraction.* We extract visual, audio and motion features that are pre-trained by different public dataset. Since VLAD aggregation layer doesn't have the ability to model temporal information, aside from the frame level features, we also extracted spatial-temporal features with 3d convolutional network and found them vital to action related class like high jump, baby crawling, etc. The details of each modality are introduced in Section 2.2.

*2.1.2 Data processing.* For the modality feature pre-processing, we use PCA, whitening and quantization. The PCA dimension for each modality is chosen according to the estimated importance to classification in common sense, for example ImageNet pre-trained features have 1024 dimension while audio feature has only 128 dimension. The whitening centralizes the energy and we clip the value to [-2.5, 2.5] followed by 8-bit uniform quantization. The purpose of quantization is to save the feature volume and the experiments show it will not hurt the performance greatly. In terms of sampling policy, we use random sampling in both training and test as illustrated in Figure 2. First we divide the video to splits with 10 minutes each so as to deal with extremely long videos. Then, we extract frame level visual feature with 1 fps and randomly select 50 frames. We found the pattern that in many classes, representative scenes are not evenly distributed. For example, "Food making" classes often start with people introducing the recipe for a long time. Evenly split videos will cause misleading train data since many scenes with "people talking" without any hints of food labeled as a particular food. Random sampling is a tradeoff between keeping key frames and computation complexity. In evaluation, we repeat the random test and average the results, it will promote the mAP about 0.1% - 0.2%. For spatial-temporal features, sampling policy applied on features not frames because each feature is influenced by nearby several frames.

*2.1.3 Feature aggregation.* We use VLAD as that in [4] to aggregate multi-modality features through time. Each modality will learn VLAD encoding and concatenate together followed by fully connect, mixture of experts and context gating.

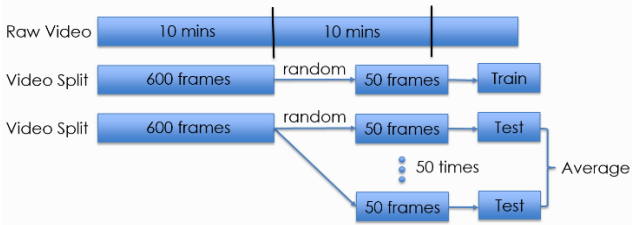

**Figure 2:** Frame level feature *Random Sampling in training and test with 1 FPS.*

## 2.2 Modality Extraction

In this section, we describe all the modalities respectively. We outline the overview of extraction in table 1.

**Table 1:** *Multi-modal Feature Extraction Overview*

| Modality | FPS | Dataset | CNN Structure |
| --- | --- | --- | --- |
| Visual | 1 | ImageNet | Inception Resnet V2 |
| Visual | 1 | ImageNet | Squeeze & Excitation |
| Visual | 1 | Places365 | Resnet152 |
| Visual | 1 | Food101 | InceptionV3 |
| I3D RGB | 0.3 | Kinetics | InceptionV1 3D |
| I3D Flow | 0.3 | Kinetics | InceptionV1 3D |
| Audio | 0.9 | AudioSet | VGG-like |

*2.2.1 Visual feature pre-trained on ImageNet.* ImageNet is a large-scale annotated dataset with 1000 categories and over 1.2 million images. CNN can learn meaningful representation after training on ImageNet. LSVC2017 provided frame level features with VGG structure. Considering VGG is not state-of-the-art CNN structure, we download 3T raw videos and extract the features on our own. We use Inception Resnet V2 [5] and Squeeze & Excitation model [6] for comparison.

*2.2.2 Visual feature pre-trained on Places365.* Places365 is the largest subset of Places2 Database [7], the 2rd generation of the Places Database by MIT CS&AI Lab. By adding the modality with this scene dataset, we hope it helps to define a context in frame level feature.

*2.2.3 Visual feature pre-trained on Food101.* In LSVC2017 dataset, about 90 classes are food related. We found food class mAP is always lower than the whole by about 15% which means it greatly impacts the performance. We look into the food class and found some classes are difficult to be distinguished visually. For examples, "making tea" vs "making mile tea", "making juice" vs "making lemonade", "making salad" vs "making sandwich". Among these classes, many ingredients are similar. To make matters worse, making food always involves scenes with people introducing the recipes. Have in mind that the clue to classify food cooking classes is so subtle, it may benefits from utilizing feature pre-trained on Food dataset. Food101 [8] has 101 food categories and 101000 images. It covers most of food classes in LSVC2017.

*2.2.4 Audio feature pre-trained on AudioSet.* Audio contains a lot of information that helps to classify videos. We extract audio feature by a VGG like acoustic model trained on AudioSet [9] which consists of 632 audio event classes and over 2 million labeled 10-second sound clips. The process is the same as that in Youtube-8M, Google has released the extraction code in tensorflow model release.

*2.2.5 Temporal-Spatial feature pre-trained on Kinetics.* Action classification is one of the hottest topics in video classification. Actions involve strong temporal dependent information that can depart action classification from single-image analysis. A lot of action dataset came up in recent years, like Kinetics [10], UCF-101, HMDB-51 etc. Action dataset has trimmed videos and each clip lasts around 10s with a single class. Carreira et al. proposed an inflated 3D model [11] that can leverage ImageNet by inflating 2D ConvNets into 3D. Their I3D model pre-trained on Kinetics gets state-of-the-art performance in both UCF101 and HMDB51 datasets. In untrimmed videos, features through time may be much more complicate, so we combine Temporal-Spatial feature I3D and Aggregation layer VLAD and the results show noteworthy improvement.

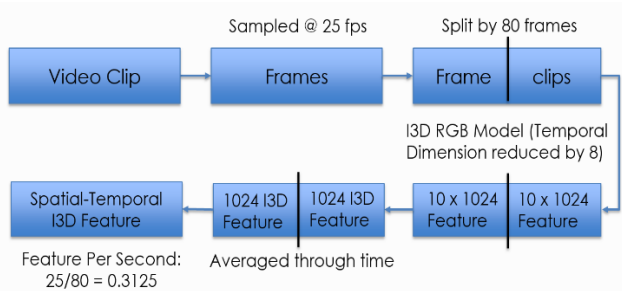

**Figure 3:** I3D RGB extraction diagram

I3D RGB feature extraction details are shown in Figure 3. For each input video clip, we first sample frames at 25 fps following the origin pre-train sampling policy and send frames to I3D model every 80 frames. Due to the 3D ConvNet structure, the temporal dimension for output feature is reduced by a factor of 8 compared with input. We averaged the output feature through time and get the Spatial-Temporal feature with FPS (Feature per second) at 0.3125. For I3D Flow, most of the part is the same except that we apply TV-L1 optical flow algorithm after sampling the videos.

In terms of realistic untrimmed videos in dataset like Youtube-8M and LSVC2017, many classes can only be distinguished by temporal information as illustrated in Figure 4. Each row shows 5 sample frames. The labels for the three videos are "baby crawling", "playing with nun chucks" and "cleaning a white board". All the videos are hard to infer ground truth based on frames. The baby could be sitting on the bed. Nun chucks are hard to notices in the second example and it seems that he is dancing. In the last video, we are not sure whether he is cleaning the board or writing on the board. VLAD and random sampling with frame level features can only aggregate single-image visual feature. Spatial-Temporal features are able to extend the learned representative feature to more complicated continuous event.

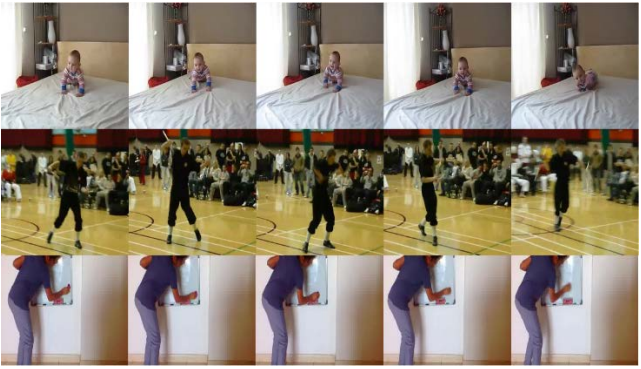

**Figure 4:** Action video frame samples in LSVC2017.

## 3 EXPERIMENT

### 3.1 Visualization

In this Section, we focus on what has been learned in VLAD and how each modality takes effect. We visualize the learned cluster and the whole aggregation process in prediction with the best single model including 5 modalities: I3D RGB, I3D Flow, Inception Resnet V2, Squeeze & excitation and food.

*3.1.1 VLAD cluster visualization.* VLAD cluster are supposed to learn meaningful visual concepts. In our implementation, we noticed that increasing the cluster size greatly doesn't improve but hurt the performance. After doing some experiments, the cluster size is set with value 40 for food, scene & audio modality and 80 for the others. We randomly picked frames in validation set and computed VLAD cluster assignment map. We illustrate some sample frames that maximize the assignment in some cluster in Figure 5.

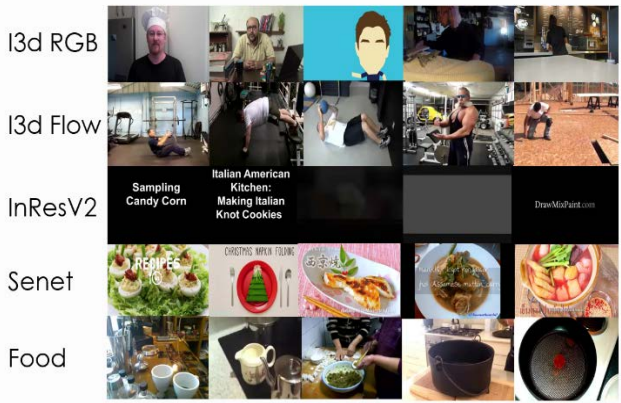

**Figure 5:** *Representative images that have largest assignment for some VLAD clusters, which successfully learn meaningful visual concept. Each row for a modality.*

*3.1.2 Aggregation visualization.* To verify the impacts with different modalities we visualize the process of aggregation. We shows the raw videos, ground truth probability changing and cluster assignment histogram in each modality. The histogram color is computed by the difference between GT probability with the one that pads the modality data with zero. The darker the histogram color is, the larger the gap is, thus the more contribution the modality makes. Different kinds of examples are shown in figure 6-8.

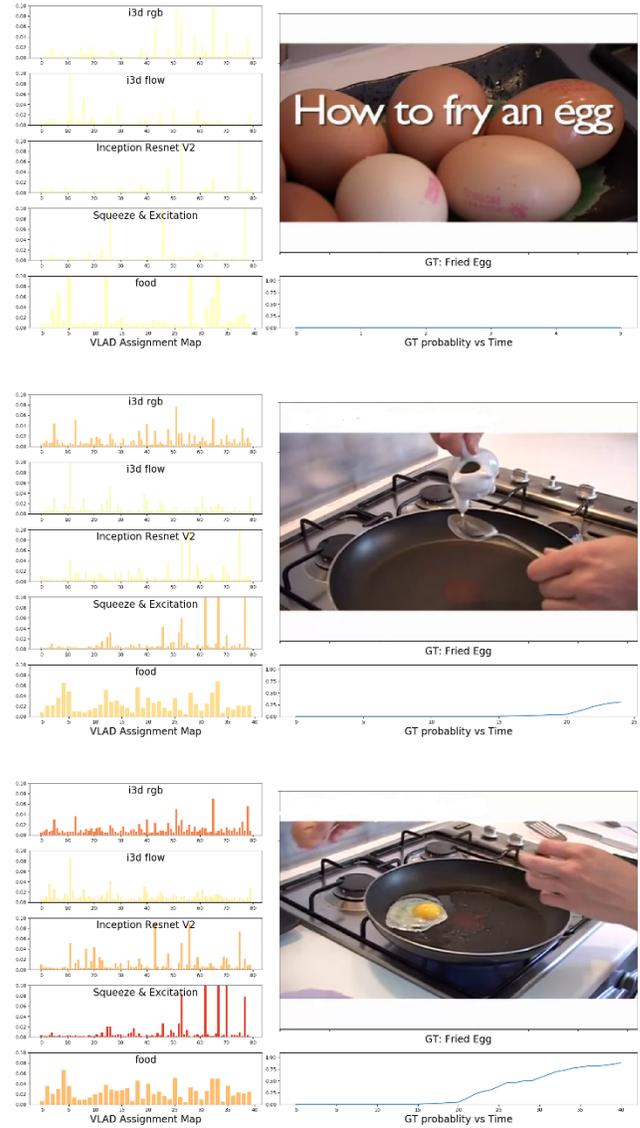

**Figure 6:** *Aggregation Visualization for class: Fried egg. Left five cluster assignment histograms are computed with I3D RGB, I3D Flow, Inception Resnet V2, Squeeze & excitation, food respectively. The top right image is a sample frame and the top bottom is the curve of the ground truth probability vs time. Here we give 3 status in the order of time. Note that in the beginning, the eggs cannot impact the probability at all. After a while, some visual hints like pouring oil and pot that highly correlated with "Fried egg" start to activate GT prediction. When the Fried egg eventually forms, it has a high confidence in GT. The histogram color shows ImageNet pre-trained feature has the most influence in this case and food/I3D RGB modality also contribute a little bit. The blue arrow in last status points to the rapid histogram change once egg changes to fried form.*

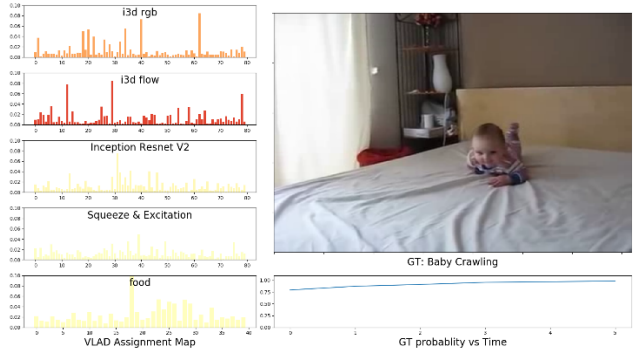

**Figure 7:** *Aggregation Visualization for class: Baby Crawling. As mention in Figure 4, this class is hard with only frame level features. The histogram color proves that only I3D features take effect.*

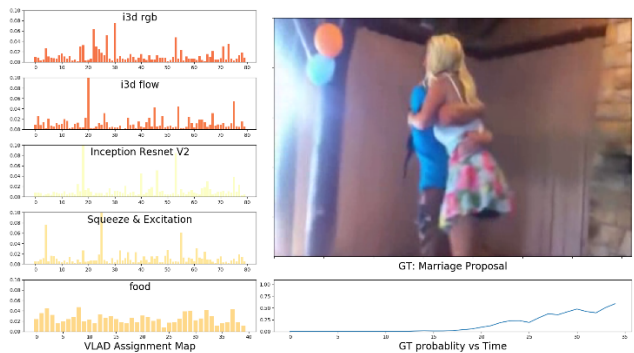

**Figure 8:** *Aggregation visualization for class: Marriage Proposal. This class has the pattern that there is always a surprise at the end. The probability curve fits well with this pattern. The value get to highest level when the couple hug each other and spatial-temporal feature successfully capture this key movement.*

### 3.2 Experiment Results

*3.1.2 Evaluation of single-modal.* We evaluate all single modality model on validation set except food because food is not a general feature for videos. Two ImageNet pre-trained modalities gets the highest mAP. CNN structure of Squeeze & Excitation is better than that of Inception Resnet V2 by nearly 3%. Spatial-Temporal feature I3D has slightly low performance. It makes sense because kinetics dataset has mainly action knowledge while LSVC2017 involves many object classes. Scene gets mAP of 0.6392 and Audio has the lowest mAP of 0.1840. Details are listed in Table 2.

**Table 2:** *Evaluation of single-modal on Validation Set*

| Modality | mAP |
| --- | --- |
| Inception Resnet V2 | 0.7551 |
| Squeeze & Excitation | 0.7844 |
| Scene | 0.6392 |
| I3D RGB | 0.7438 |
| I3D Flow | 0.6819 |
| Audio | 0.1840 |

**Table 3:** *Evaluation of multi-modal on Validation Set*

| Multi-modality | mAP | mAP(food) |
| --- | --- | --- |
| I3D | 0.7890 | 0.5309 |
| I3D + InResV2 | 0.8130 | 0.6070 |
| I3D + InResV2 + Audio | 0.8373 | 0.6557 |
| I3D + InResV2 + Food | 0.8246 | 0.6710 |
| I3D + Senet | 0.8395 | 0.6652 |
| I3D + Senet + Food | 0.8428 | 0.6855 |
| I3D + Senet + Scene | 0.8379 | 0.6670 |
| I3D + Senet + InResV2 | 0.8449 | 0.6901 |
| I3D + Senet + InResV2 + Food | 0.8485 | 0.7017 |
| 25 model ensemble | 0.8848 | 0.7478 |
| 25 model ensemble (on Test) | 0.8741 | unknown |

*3.1.2 Evaluation of multi-modal.* In Table 3, we shows the multi-modality model results. I3D RGB and Flow are default modalities. By comparing I3D with I3D + Senet and Senet in Table 2 it is clear that spatial-temporal feature pre-trained on action dataset and ImageNet pre-trained frame level features complement each other well, the combination gets a relative high mAP of 0.8395. By adding more modalities based on I3D and Senet, the best multi-modal single model achieves mAP of 0.8485. Since food is a very important subset, we list mAP of food in the third column, it proves that food modality helps the food performance by a considerable margin. Audio can improve the mAP while scene seems to be useless in our results. Our final submit is an ensemble of 25 models with different combination of modalities. It gets mAP of 0.8741 on test and ranked 1st in the competition.

## 4 CONCLUSIONS

In summary, we have proposed a multi-modal aggregation method for large-scale video classification. We showed that spatial-temporal features pre-trained on action dataset improves the performance a lot. We also visualize the aggregation process and find that multi-modalities are mutually complementary and the model implicitly selects the modality that best describe the videos.